\documentclass[runningheads]{llncs}
\usepackage[T1]{fontenc}
\usepackage{amsmath} 
\usepackage{amssymb}  
\usepackage{graphicx}
\usepackage{diagbox}
\usepackage{xcolor}
\usepackage{cite} 
\begin{document}
\title{Sample-efficient Low-level Motion Planning for Robotic Manipulation Tasks via Zero-shot Transfer Learning}
\titlerunning{Sample-efficient Low-level Motion Planning via Zero-shot Transfer Learning}
%
%
\author{%
Yuanzhi He\inst{}\thanks{Corresponding author.} \and
Victor Romero-Cano \and
José J. Patiño \and
Juan David Hernández \and
William Sawtell \and
Gualtiero Colombo%
}

\institute{School of Computer Science \& Informatics, Cardiff University, Cardiff, UK\\
\email{HeY65@cardiff.ac.uk}}
\authorrunning{Y. He et al.}

%
%
%
\maketitle              
\footnotetext[1]{The code is open-sourced at https://github.com/YuanzhiHe/iCEM\_TL.git.}
\footnotetext[2]{This paper has been accepted to ICANN 2026.}
\begin{abstract}
As robotic systems become more sophisticated, the growing complexity of their motion planning models and the longer training times pose substantial challenges. Evolutionary algorithms such as the Sample-efficient Cross-Entropy Method (iCEM) have recently demonstrated promising potential for low-level real-time planning by leveraging efficient knowledge reuse strategies to improve performance. Although effective in many control tasks, iCEM's performance can be constrained in more complex scenarios, particularly those requiring stacking, sliding, and shelf placement. In this work, we propose a novel iCEM+TL framework that explicitly leverages Transfer Learning (TL), where key iCEM parameters are transferred from simpler upstream tasks to guide more complex downstream tasks. Additionally, we applied Reward Redesign (RR) through task decomposition for stacking objects and shelf placement to optimize task-specific performance. Results from the simulation show that our framework achieves success rate improvements of up to $23\%$. The framework is further validated on a real Franka Emika robot in a stacking task, demonstrating its practical feasibility for real-world deployment.

\end{abstract}

\section{Introduction}

In recent years, research on robotic manipulation has increasingly focused on advancing from basic tasks such as pushing, reaching, and placing objects to more complex challenges, including sliding objects, stacking multiple items, and interacting with external structures, for example, picking and shelf placement.

High-level learned models such as Deep Reinforcement Learning (DRL) are able to achieve reliable performance, although requiring significant computational resources and extended training times~\cite{soori2023artificial, Yang2021}. To address these limitations, low-level planners based on Evolutionary Strategies (ES) such as the Sample-efficient Cross-Entropy Method (iCEM) have been proposed and are often combined with high-level planners such as learned models, to enhance overall performance and integrate planning across both macro and micro scales~\cite{Sancaktar2022}. Nevertheless, when facing increasingly complex manipulation problems, low-level planners are not sufficient, even when combined with high-level motion planners.


Task decomposition has emerged as an effective strategy to improve scalability and generalization~\cite{jaquier2025transfer, zhu2023transfer}. Two complementary techniques are particularly relevant: Transfer Learning (TL) and Reward Redesign (RR). TL reuses knowledge like the trained parameters acquired from upstream (simpler) tasks to accelerate learning in downstream (more complex) tasks, thereby reducing retraining requirements and improving efficiency. In addition, RR refers to tailoring reward structures to better align with task-specific objectives and guide exploration. Existing research favouring TL primarily emphasizes either validating the effectiveness of transfer for learned models, such as DRL algorithms through Single and Double Transfer~\cite{He2024}, or exploring the advantages of training models via a sequence of progressively complex tasks, a strategy often referred to as goal-oriented computing~\cite{Yang2021, colas2019curious}. 

Therefore, in this study we propose a novel framework iCEM+TL, for task decomposition that integrates TL with RR to enhance the inference process on robotic manipulation tasks. Specifically, we optimize the original iCEM algorithm to incorporate TL and RR in order to provide clearer guidance for the downstream stacking, sliding, and shelf placement tasks. This results in an improved performance on all the experimental simulated environments.

There are two primary contributions to this work. Firstly, by utilizing task decomposition principles, we improve performance on more challenging robotic manipulation tasks, including stacking, sliding, and shelf placement, using only a low-level motion planner without relying on high-level learned models. Secondly, we investigate the modalities of the application of task decomposition in the context of ES. Specifically, we provide more insights on the transferable components (‘what to transfer’) and strategies (‘how to transfer’) during TL and how the reward functions can be decomposed into more elementary modules in the RR phase. As a result, we prove the ability of our iCEM+TL framework to lead the robot to conduct simulation tasks including stacking, sliding and shelf placement. Finally, a real-world validation for the Stack task is deployed to further prove the feasibility of our framework.


\section{Related Works}

DRL has achieved strong performance in robotic manipulation, although it often requires large amounts of interaction data and long training times. Soft Actor-Critic (SAC)~\cite{haarnoja2018soft} improves exploration through entropy maximization, while Truncated Quantile Critics (TQC)~\cite{kuznetsov2020tqc} further enhances stability by reducing overestimation bias using distributional value truncation. However, DRL methods typically struggle in sparse-reward settings such as FetchSlide, where exploration becomes inefficient. To address this issue, Hindsight Experience Replay (HER) relabels failed trajectories with achieved goals to improve sample efficiency~\cite{andrychowicz2017hindsight}. While HER is effective for simpler tasks such as sliding, it remains less effective in more complex scenarios like stacking and shelf placement, where long-horizon dependencies and sparse rewards increase task difficulty.

Evolutionary algorithms provide an alternative paradigm that avoids gradient-based optimization and instead searches directly in the action space. Structured exploration strategies have been shown to improve performance in difficult exploration problems~\cite{Ecoffet2021}. Among these approaches, ES are particularly effective for optimizing high-dimensional control policies~\cite{Salimans2017}. A representative method is the Cross-Entropy Method (CEM), which samples candidate trajectories and iteratively refines the sampling distribution based on elite solutions. The improved variant iCEM~\cite{Pinneri2021} introduces colored noise and elite reuse across timesteps, significantly improving sampling efficiency. Related work such as CEE-US~\cite{colas2019curious,Sancaktar2022} further integrates iCEM with intrinsic motivation for multi-goal tasks.

Beyond optimization-based approaches, recent studies have explored generative imitation learning methods for low-level control, where policies are learned by explicitly modeling the distribution of expert trajectories. For example, PointFlowMatch~\cite{chisari2024pointflowmatch} employs conditional flow matching, a flow-based generative model that learns a continuous transformation from a simple prior distribution to a complex trajectory distribution conditioned on point cloud observations, enabling multimodal action generation. 
Recent foundation models have also shown strong performance in high-level robotic planning and perception. 
For instance, Neural MP~\cite{dalal2024neuralmp} trains a generalist neural motion planner through large-scale procedural scene generation and expert distillation, achieving strong cross-scene generalization. 
However, these approaches focus on joint-space trajectories and do not address grasping policies required for full manipulation.

TL has been widely studied as a strategy for improving sample efficiency and cross-task generalization in robotics. 
Recent work has investigated task-to-task transfer, domain adaptation, and zero-shot transfer settings~\cite{jaquier2025transfer,zhu2023transfer}. 
In manipulation tasks, few-shot transfer methods have demonstrated that reusing policy parameters can significantly accelerate learning and improve performance~\cite{He2024}.

In contrast to these approaches, our work focuses on improving low-level motion planning through task decomposition and knowledge reuse within an optimization-based planner. Rather than relying on long training processes or learning large neural policies, our framework operates directly at planning time by extending the iCEM algorithm. Specifically, TL is incorporated by reusing sampling distributions and elite trajectories from upstream tasks, while RR decomposes complex manipulation objectives into structured sub-goals. This integration improves the efficiency of trajectory search and leads to better performance on structured manipulation tasks such as stacking and shelf placement.

\section{Proposed Approach}

\subsection{Problem Formulation}

Increasingly complex robotic manipulation tasks, including multi-object stacking, object rearrangement, and assembly operations, require precise and efficient action sequences.
To systematically address these tasks, we firstly define each task by a set of objects $\mathcal{O}$, an initial configuration $s_0$, and a goal configuration $s_g$. Here, $\mathcal{O}$ is implicitly encoded in the state representation $s_t$. The objective is to find a sequence of actions $a_{0:T} = \{a_0, a_1, ..., a_T\}$ that optimally moves the objects in $\mathcal{O}$ from $s_0$ to $s_g$, according to the modalities of the specific task.

Formally, this can be formulated as an optimization problem over trajectories that maximize the expected cumulative reward:

\begin{equation}
    \max_{a_{0:T}} \sum_{t=0}^{T} R(s_t, a_t),
\end{equation}

where $R(s_t, a_t)$ is a task-specific reward function.

\subsection{iCEM+TL Framework}
To tackle the challenges of overlong training and inefficient and under-performing sampling, we propose a model-based planning framework that integrates iCEM with TL and RR.
\begin{figure}[t]
    \centering
    \includegraphics[width=0.9\textwidth]{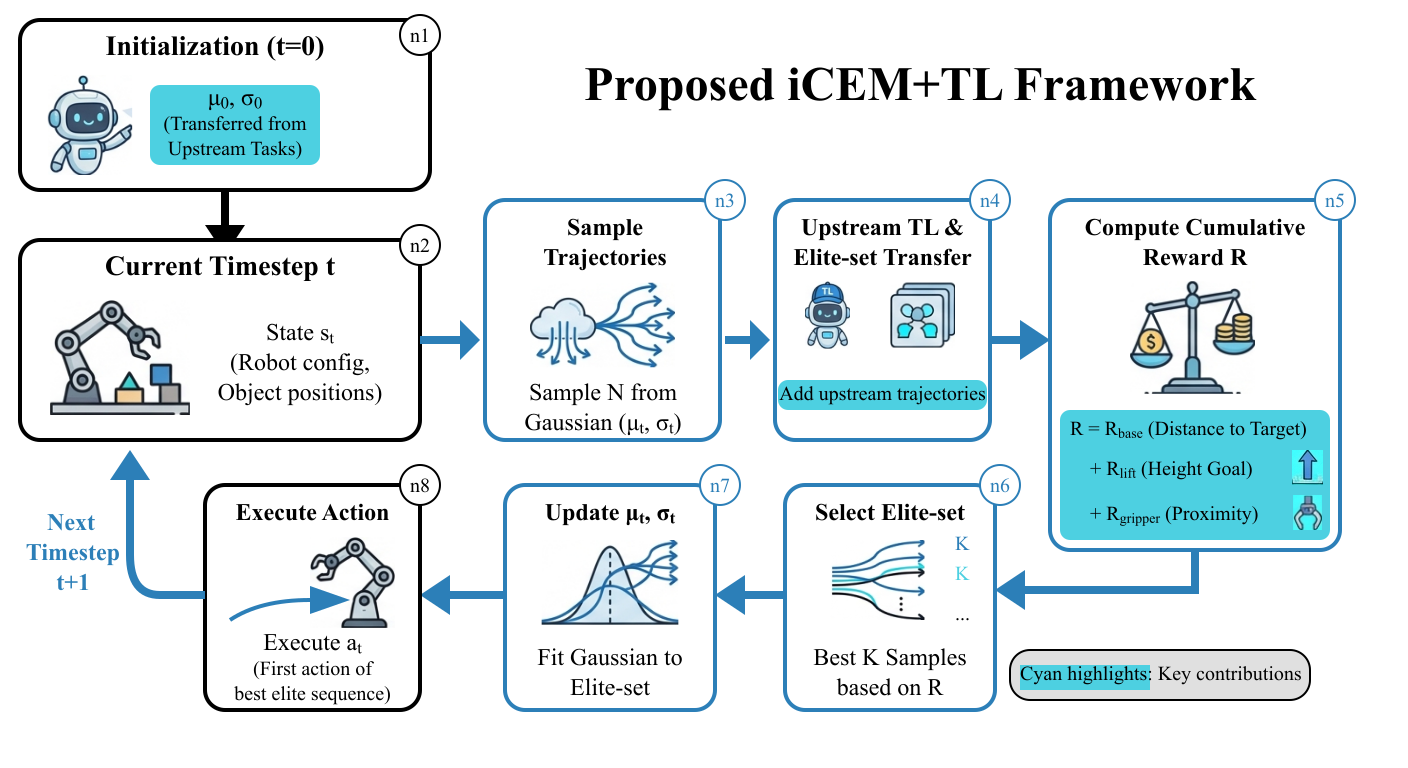}
    \caption{Diagram of our iCEM+TL framework showing at timestep $t$ the last iteration of our inner loop ($i=I-1$) with the output of the action $a_t$. Color blue is the additional processes of our proposed method integrating iCEM with TL and RR.}
    \label{fig} 
\end{figure}

As shown in Fig.~\ref{fig}, our framework operates as follows: at timestep $t=0$, we transfer the parameters $\mu$ and $\sigma$, which denote the mean and standard deviation of the Gaussian sampling distribution in iCEM. These parameters are estimated from the elite trajectory distribution of the upstream task, providing a more informative initial sampling distribution for the downstream task. This transfer enables better-guided exploration by initializing the sampling process closer to promising regions of the action space. Here, zero-shot means that no downstream training is required: the solution is still obtained through online iCEM optimization, with the transferred components only initializing and guiding the search. Furthermore, at every timestep $t$, in addition to keeping and shifting elite trajectories from the previous timestep $t-1$ and sampling candidate trajectories following the original iCEM procedure, we transfer and add the best-performing trajectories from the upstream task to the elite trajectory set, thus providing a level of guidance to the current task. In detail, candidate trajectories are sampled with a Gaussian distribution, and during optimization the elites are selected according to their cumulative reward over the trajectory horizon $h$:
\begin{equation}
    G(\mathbf{a}_{0:h-1}) \;=\; \sum_{t=0}^{h-1} r(s_t, a_t),
\end{equation}
where $r(s_t,a_t)$ denotes the step-wise reward obtained by executing action $a_t$ in state $s_t$. $G(\mathbf{a}_{0:h-1})$ then represents the cumulative reward of a trajectory within the horizon $h$, with higher values preferred. The distribution parameters are then updated based on the selected elites, with momentum to smooth the updates across iterations. 
For each timestep $t$, the planner generates a sample, elite set and fits the Gaussian distribution to the new mean $\mu$ and standard deviation $\sigma$ for a number of iterations.
After the final iteration, the first action of the best elite sequence is executed, and the process repeats at the next timestep.
By reusing related experience (e.g., elite trajectories and sampling distribution attributes) and selecting the best trajectory at each timestep (the one with the highest cumulative reward), the planner consistently executes higher-quality actions, as evidenced by the improved performance observed in our experimental results. The complete pseudocode of the proposed iCEM+TL framework is provided in our open-sourced repository.\footnotemark[1]

\begin{figure}[t]
    \centering
    \includegraphics[width=80pt, height=80pt]{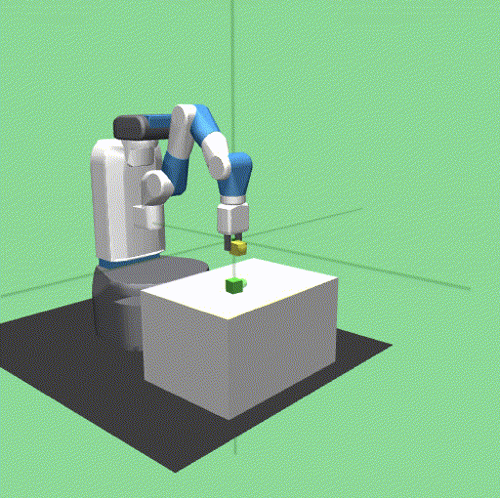}
    \includegraphics[width=80pt, height=80pt]{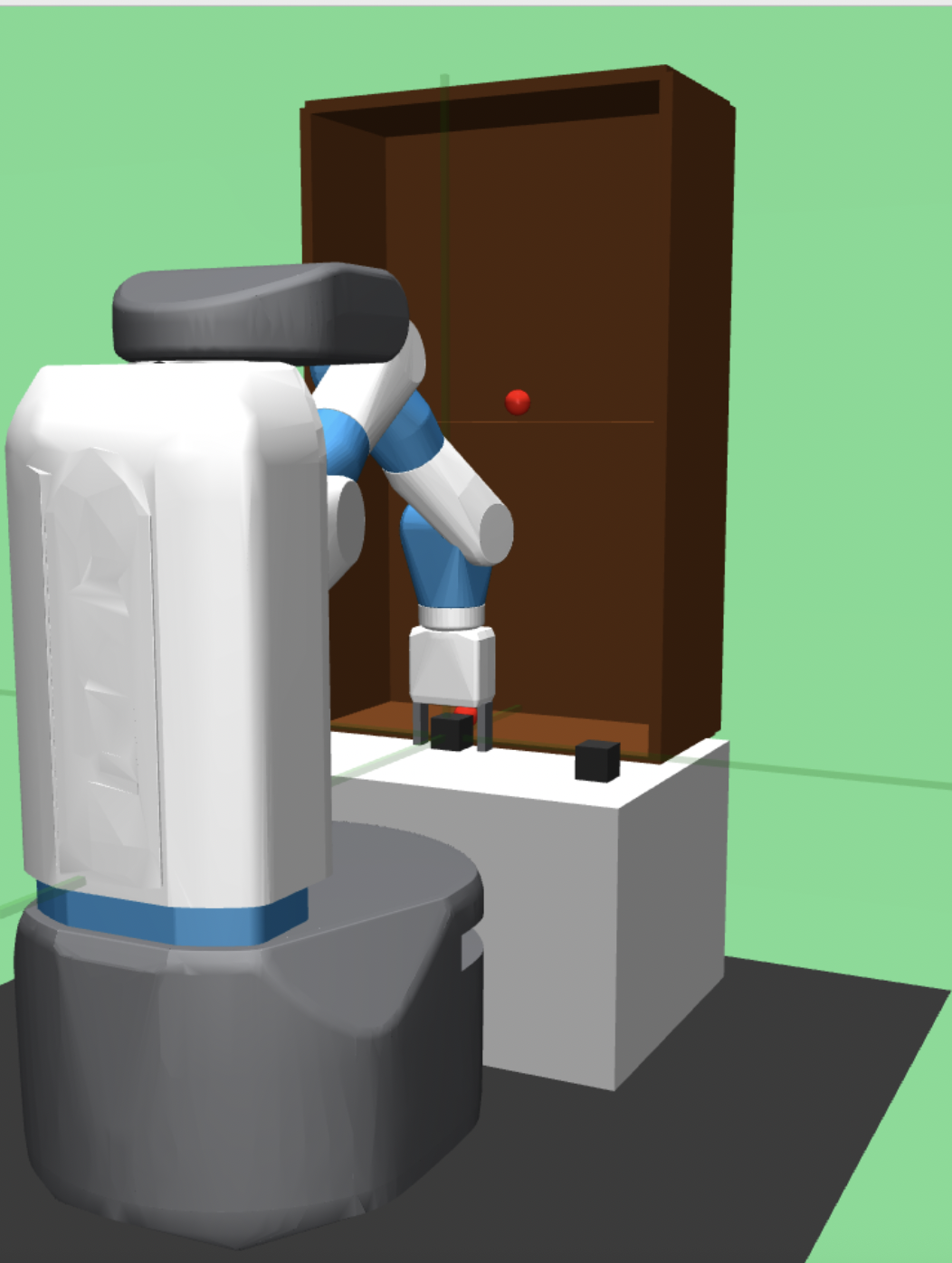}
    \includegraphics[width=80pt, height=80pt]{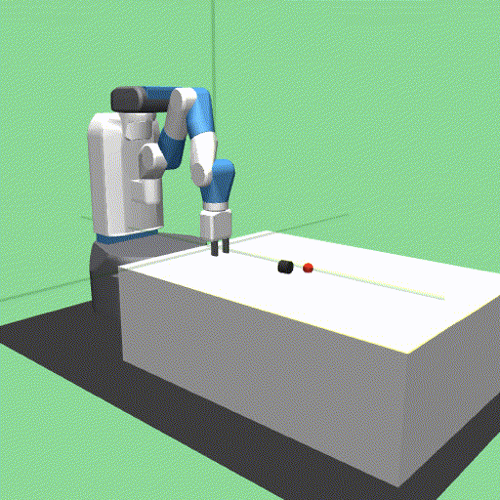}
    \caption{FetchStack (left), Shelf (middle) and FetchSlide (right) tasks in MuJoCo simulation} 
    \label{fetch_stack_slide_and_shelf}
\end{figure}

To better handle the complexity of the Stack and Shelf tasks, we decompose the task objective into simpler sub-goals and redesign the reward as

\begin{equation}
    R = R_{base} + R_{lift} + R_{gripper},
\end{equation}

where the base reward encourages accurate object placement:

\begin{equation}
    R_{base} = -d_0 + \mathbb{1}_{near} \cdot (-d_1),
\end{equation}

where $d_0 = |p_{obj0} - p_{obj0_{target}}|$ and $d_1 = |p_{obj1} - p_{obj1_{target}}|$ denote the distances between objects and their targets. 
The indicator $\mathbb{1}_{near}$ activates the second term once the lower object is sufficiently close to its goal.

The gripper reward encourages the gripper to remain close to the manipulated object:

\begin{equation}
    R_{gripper} = -|p_{gripper} - p_{obj}|,
\end{equation}

where $p_{obj}$ denotes the currently active object.

Finally, the lifting reward promotes correct vertical motion during stacking:

\begin{equation}
    R_{lift} =
    \mathbb{1}_{near}
    \left(
    -0.5 |h_{obj1} - h_{target}|
    + 0.8 \max(h_{obj1} - h_{obj0}, 0)
    \right).
\end{equation}

where $h_{obj1}$ and $h_{obj0}$ are the heights of the upper and lower objects, and $h_{target}$ is the target stacking height.

We further propose a principled criterion to assess whether an upstream task is expected to provide beneficial transfer. Intuitively, not all upstream tasks are equally useful, and effective transfer requires structural alignment between upstream and downstream tasks. To quantify this, we evaluate whether trajectories from the upstream elite set remain high-quality under the downstream reward.

For a trajectory $\tau$, the downstream reward is defined as
\begin{equation}
R_D(\tau) = \sum_{t=0}^{h-1} r_D(s_t, a_t),
\end{equation}
and we measure the elite reward advantage as
\begin{equation}
\label{eq:elite_alignment}
\Delta R_{U \rightarrow D} =
\mathbb{E}_{\tau \sim \mathcal{E}_U} [R_D(\tau)]
-
\mathbb{E}_{\tau \sim \mathcal{B}} [R_D(\tau)],
\end{equation}
where $\mathcal{E}_U$ denotes the elite set from the upstream task and $\mathcal{B}$ denotes trajectories sampled from the initial downstream distribution. A positive $\Delta R_{U \rightarrow D}$ indicates that upstream elites already achieve higher rewards under the downstream objective than baseline samples, suggesting that the upstream task is well-aligned with the downstream task and can provide effective guidance for trajectory search.

\section{Experimental Results and Discussion}

In this section, we show the result of the simulation concerning a number of robotic manipulation tasks such as Fetch Stack, Slide and a newly generated Shelf environment, see Fig. \ref{fetch_stack_slide_and_shelf}. 
All the simulation experiments are run on the MuJoCo Fetch simulation. The experimental environment FetchSlide is chosen from the MuJoCo manipulation Fetch environments provided by OpenAI Gym~\cite{brockman2016openai}. The FetchStack environment is adapted from an existing version that was previously used to evaluate a GNN-ensemble combined with the iCEM method in~\cite{Sancaktar2022}. In our version, we modified the logic for object placement and goal generation to better suit our experiments. The goal of the FetchStack task is to first pick an object from a source point and place it at a target point, and then pick the other objects and stack them in order on the first object. Furthermore, we created a unique Shelf environment, where the robot has to pick a number of objects one by one and place them on different layers. There is also a single-object version of the task, which serves as the upstream task for transfer, and consists of picking one object and placing it on a random shelf layer. The experiments were repeated over three runs to reduce the randomness inherent in iCEM, and we report average success rates with standard deviations. To enhance exploration and allow the transfer of trajectories from upstream tasks, we set the elite set size to 20, thus resulting in 6 samples shifted from the previous timestep, 10 transferred from an upstream task, and 4 generated through Gaussian sampling. The experiments were performed on a PC with an Intel Core i7-14700K CPU and a supercomputer with an Intel(R) Xeon(R) Gold 6148 CPU.

\begin{figure*}[t]
    \centering
    \includegraphics[width=0.84\textwidth]{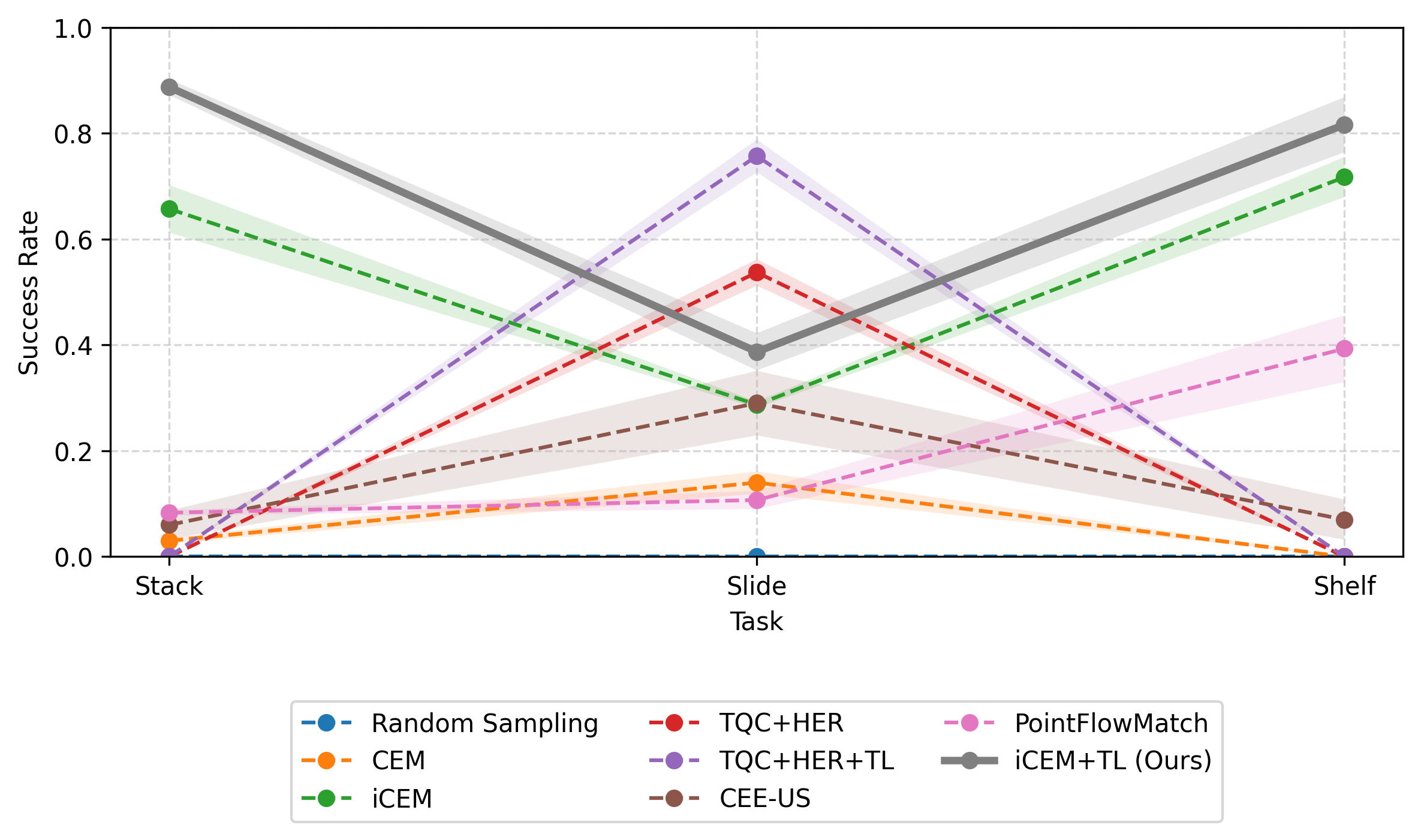}
    \caption{Average success rate comparison between baselines (Random Sampling~\cite{Pinneri2021}, CEM~\cite{Pinneri2021}, iCEM~\cite{Pinneri2021}, TQC+HER~\cite{kuznetsov2020tqc}, TQC+HER+TL~\cite{He2024}, CEE-US~\cite{Sancaktar2022} and PointFlowMatch~\cite{chisari2024pointflowmatch}) and our proposed iCEM+TL framework with RR on Stack, Slide and Shelf tasks. Standard deviation values are represented as shadows. Planning horizon $H_{Slide}$=50, $H_{Stack}$ and $H_{Shelf}$=1000. Sample Size=40. Elite Size=20} 
    \label{main_result}
\end{figure*}

Fig.~\ref{main_result} presents a comparison of success rates across three representative robotic manipulation tasks with distinct structural properties: FetchStack, FetchSlide, and the newly introduced Shelf environment. These tasks span increasing levels of difficulty, providing a comprehensive evaluation of low-level planning performance. For the PointFlowMatch baseline, we generate task-specific expert demonstrations to build the training dataset, since the original work does not provide demonstrations for the tasks considered in our environments. The collected demonstrations are subsequently used to train the PointFlowMatch model to reproduce trajectories in our task settings. The model is trained under two different epoch settings (500 and 1500), among which the 500-epoch configuration achieves superior performance and is thus reported. We note that PointFlowMatch is an imitation-based generative policy rather than an online trajectory optimizer, and its performance in our setting is likely constrained by the limited ability of offline trajectory generation to recover from compounding errors in long-horizon, contact-rich manipulation tasks.

On the Stack task, iCEM+TL significantly improves performance over the best-performing baseline iCEM, achieving an absolute gain of approximately $23.0\%$ with reduced variance. In comparison, learning-based planners exhibit clear limitations in this long-horizon setting, with TQC+HER not achieving competitive performance and PointFlowMatch performing substantially worse. On the Slide task, learning-based baselines perform more competitively, with TQC+HER and its transfer variant achieving stronger performance, while our approach remains effective without task-specific training. For the Shelf task, iCEM+TL again improves performance over vanilla iCEM, whereas learning-based methods show limited generalization, with CEE-US and PointFlowMatch remaining substantially below iCEM+TL. Overall, these results demonstrate the advantage of incorporating TL and RR within an optimization-based planning framework, particularly for long-horizon manipulation tasks.

\subsection{Ablation Study} 

\begin{figure*}[t]
    \centering
    \includegraphics[width=\textwidth]{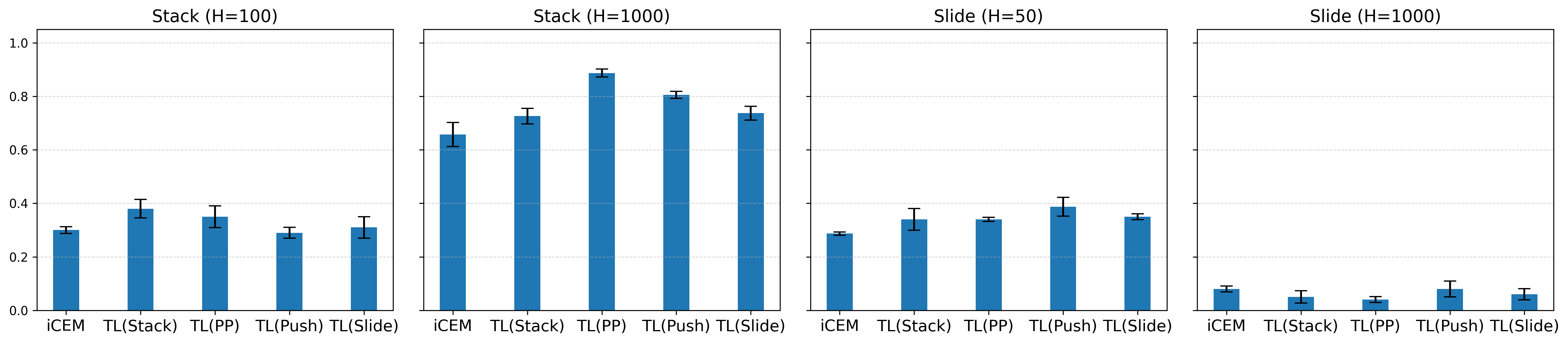}
    \caption{Average success rate comparison between iCEM and iCEM+TL (transfer from Stack, Pick\&Place, Push or Slide tasks) frameworks on Stack and Slide tasks with different Timesteps (planning horizon) with standard deviations. Sample Size=40. Elite Size=20} 
    \label{Ablation1}
\end{figure*}

An additional contribution of our study is to identify which subsets of hyperparameters from the original iCEM can be directly adopted or adapted, and how they impact different tasks. To verify the effect of each component, we conduct ablations by varying several settings. Fig.~\ref{Ablation1} compares iCEM from scratch with TL from different upstream tasks, including Stack, Pick\&Place, Push, and Slide. For each task, the planning horizon $H$ is selected to achieve the best performance, with more complex tasks requiring longer horizons.
We observe that transferring upstream Pick\&Place trajectories provides the largest improvement on Stack ($23.0\%, H=1000$). Similarly, Slide ($H=50$) improves by approximately $10.0\%$ when transferring from Push. For the Shelf task ($H=1000$), since the one-object version can be fully solved with a 100\% success rate without the aid of any TL strategy, we have therefore implemented a two-object version by transferring the best trajectories directly from the one-object version. This achieves a $9.9\%$ improvement on a version without implementing any task decomposition (raising from $0.717\pm0.038$ to $0.816\pm0.052$).


Importantly, the empirical results corroborate the proposed criterion for evaluating which upstream tasks are likely to yield positive transfer to a given downstream task. Specifically, we observe that upstream tasks with higher reward-component overlap and positive elite reward alignment ($\Delta R_{U \rightarrow D} > 0$) consistently produce larger downstream improvements. 
This is reflected in our experimental findings, where transferring from Pick\&Place to Stack and from Push to Slide leads to significant performance gains, while less related task pairs (e.g., Slide to Stack) provide limited or no improvement. These results demonstrate that effective transfer depends on structural alignment between tasks, supporting the proposed task-selection criterion.

\begin{figure*}[t]
    \centering
    \includegraphics[width=\textwidth]{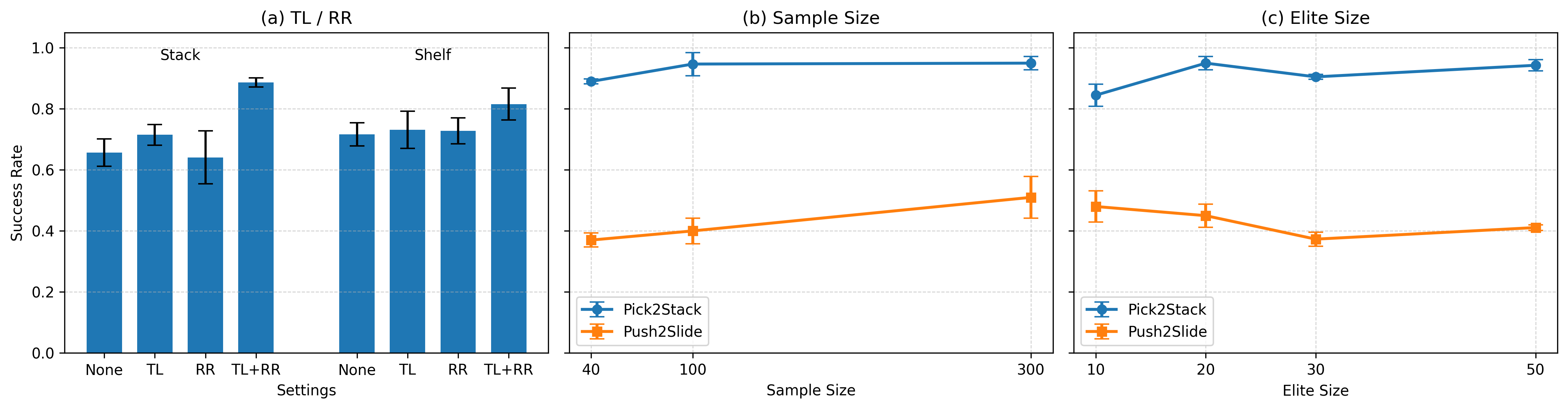}
    \caption{Average success rate comparison between iCEM and iCEM+TL with standard deviation values. (a) Enabling and disabling task decomposition features (TL \& RR) for the Stack and Shelf tasks. (b) Different sample size settings with iCEM+TL framework on Stack task (transfer from Pick\&Place) with 1000 Timesteps and Slide task (transfer from Push) with 50 Timesteps. Elite Size=20. (c) Different elite size settings with iCEM+TL framework on Stack task (transfer from Pick\&Place) with 1000 Timestep and 100 Sample Size and Slide task (transfer from Push) with 50 timestep and 300 Sample Size.} 
    \label{Ablation2}
\end{figure*}

As shown in Fig.~\ref{Ablation2}(a), we analyze the individual and combined effects of TL and RR on the Stack and Shelf tasks.
For the Stack task, applying TL alone yields a moderate improvement of approximately $5.8\%$, while RR alone slightly degrades performance by $1.6\%$. 
When both TL and RR are enabled, the success rate increases significantly, achieving an overall improvement of about $23.0\%$ over the baseline. This indicates a strong synergistic effect between TL and RR that cannot be achieved by either component individually.
A similar but less pronounced trend is observed in the Shelf task.
TL and RR individually produce only marginal improvements (around $1\%$), whereas their combination leads to a notable performance gain of approximately $9.9\%$. These results suggest that, although each component alone contributes limited improvements on Shelf, their combination is important for handling its structured and sequential nature.

We further evaluate different sample sizes (40, 100, and 300) for the Pick-to-Stack and Push-to-Slide tasks using the best upstream transfers identified in Fig.~\ref{Ablation2}(b). We observe that a sample size of 300 consistently achieves the best performance, yielding improvements of approximately $6.3\%$ for Pick-to-Stack and $14\%$ for Push-to-Slide compared to smaller sampling settings. This suggests that a moderately larger sample size improves performance by enabling broader exploration while remaining guided by the sampling distribution. Combined with the $23.0\%$ gain from TL and RR, this results in a total improvement of approximately $29.3\%$ over the vanilla iCEM baseline on the Stack task. 
However, further increasing the sample size leads to diminishing returns: larger sample sizes significantly increase inference time for long-horizon tasks such as Stack (1000 timesteps), and can degrade performance in shorter-horizon tasks such as Slide due to increased variance and reduced influence of high-quality trajectories during distribution updates.

Finally, Fig.~\ref{Ablation2}(c) analyzes the effect of the elite size on performance. We observe that a relatively small elite set yields the best results, with an elite size of 20 for Pick-to-Stack and 10 for Push-to-Slide. This indicates that a compact elite set is sufficient to effectively guide the optimization process. In all experiments, a fixed set of 10 elite trajectories is transferred from the upstream task.

These findings further support the proposed criterion for upstream task selection. While TL improves performance when task structures are aligned, it does not always yield benefits. For example, transferring from Slide to Stack does not improve performance much over vanilla iCEM, as Slide involves planar motion whereas Stack requires vertical and height-sensitive control. This mismatch leads to uninformative transferred trajectories and neutral transfer, indicating that effective transfer requires strong alignment between upstream and downstream objectives.


\subsection{Real-world Experiment}

\begin{figure}[t]
    \centering
    \includegraphics[width=0.32\textwidth]{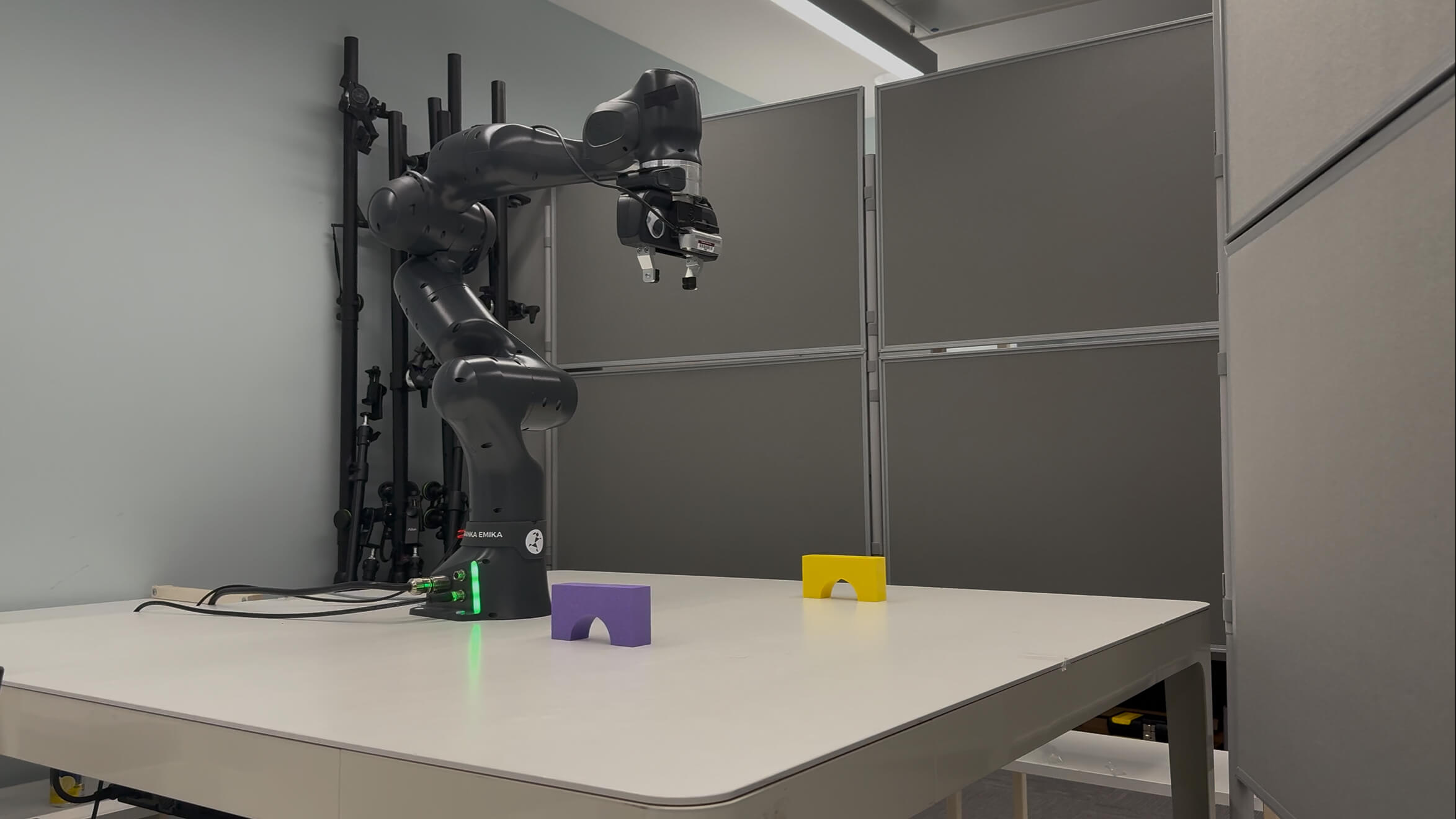}
    \includegraphics[width=0.32\textwidth]{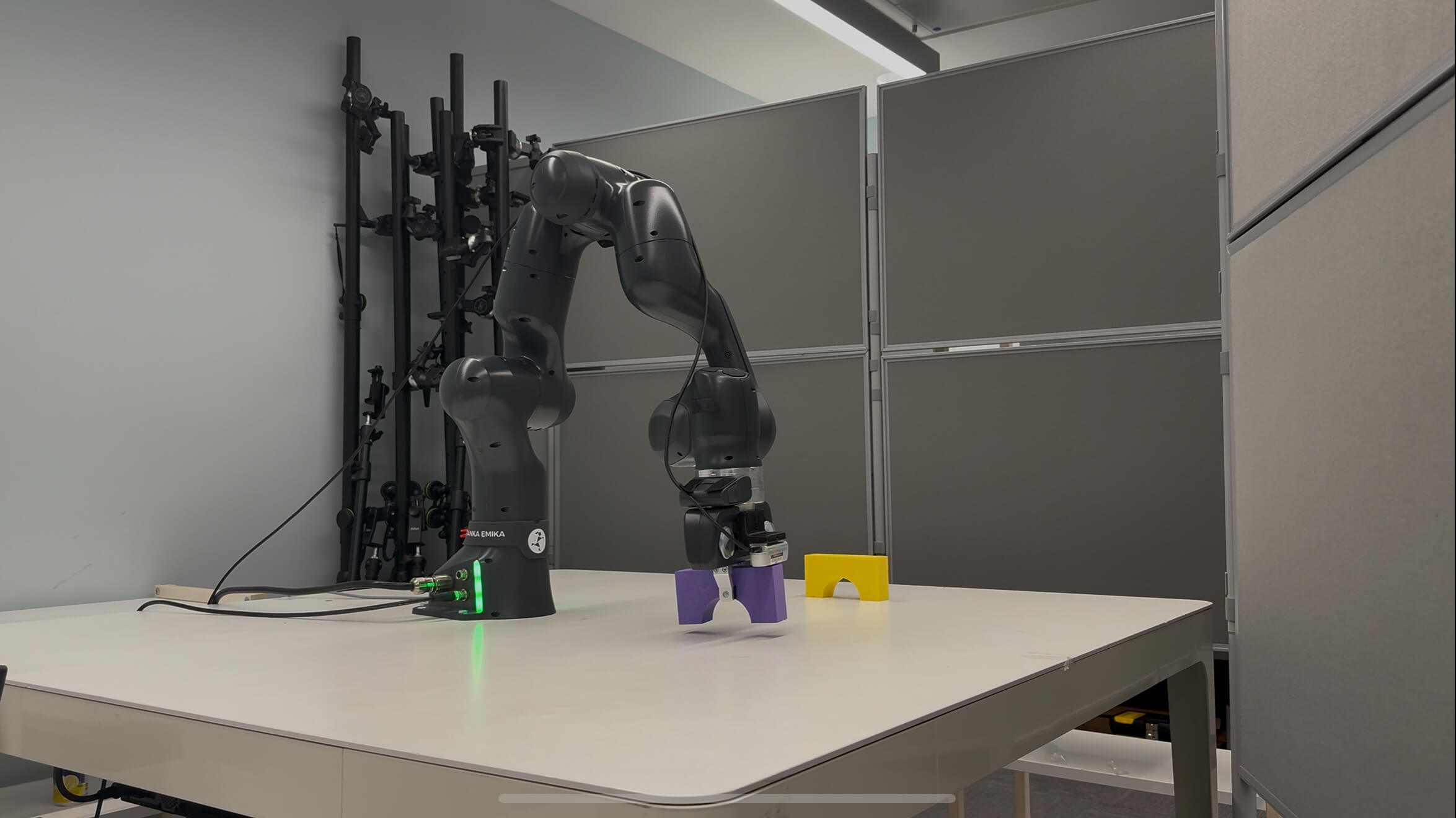}
    \includegraphics[width=0.32\textwidth]{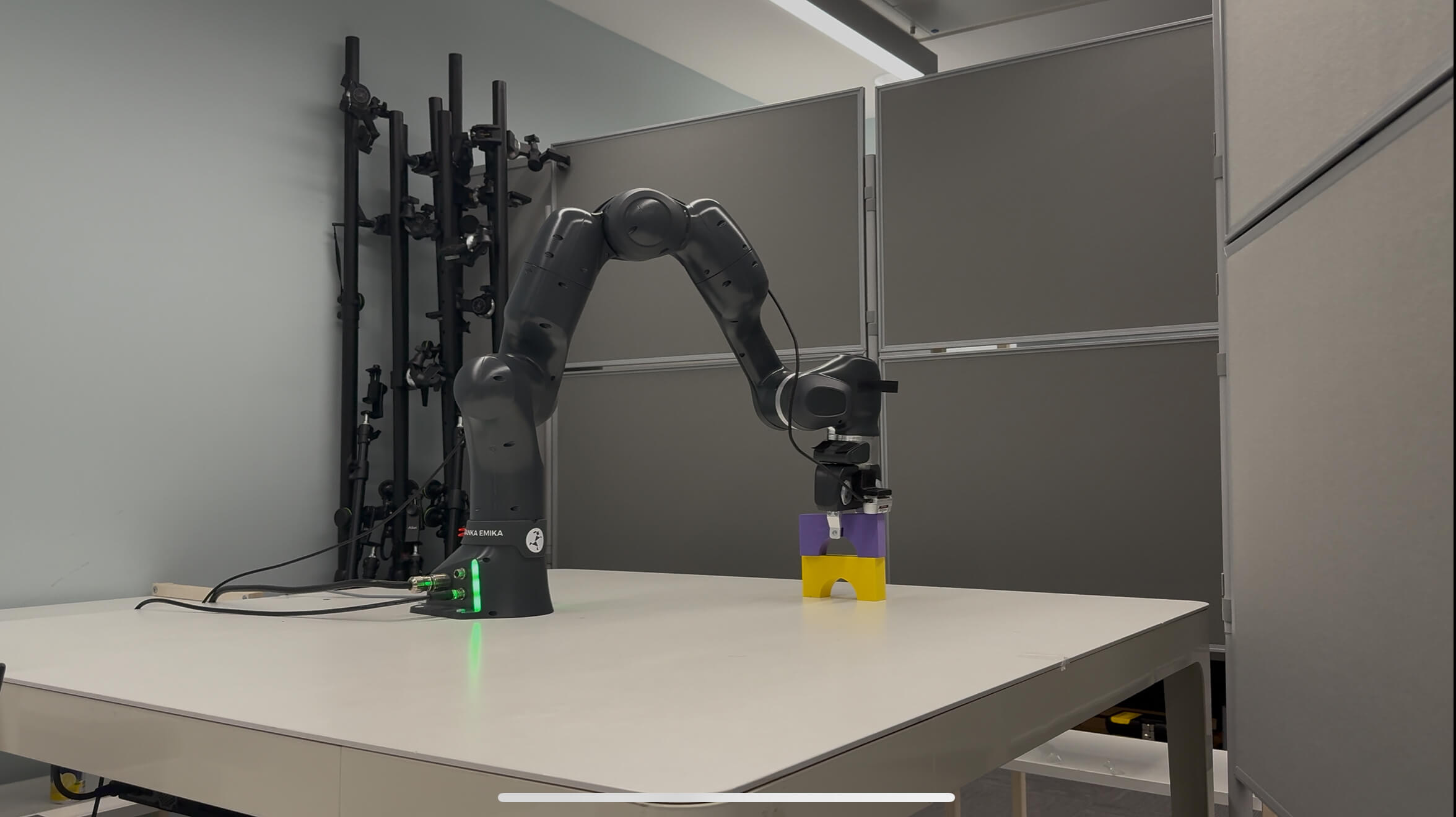}
    \caption{Real-world experiments on the stacking task with Franka Emika FR3 robot.} 
    \label{real_stack}
\end{figure}

We evaluate our iCEM+TL framework on a Franka Emika FR3 robot to demonstrate its real-world applicability for selected manipulation tasks. To ensure safe execution in real-world settings, we first build a specific simulated environment in MuJoCo that mirrors the initial end-effector and object configurations, including positions and sizes. The best-performing trajectory is saved from simulation and then executed on the real robot. We use a 4-DoF robotic arm in simulation to enable the comparison against the original iCEM and deployed on the FR3 robot with 7-DoF. Each trajectory consists of 4D actions representing the end-effector's xyz movement and gripper open/close commands. The RelaxedIK function is applied to generate multiple continuous joint solutions for each end-effector target. To account for uncertainties in trajectory interval lengths, we apply linear interpolation for the largest joint movements between steps, ensuring compliance with the robot's constraints. The real-world execution demonstrates that our framework can effectively handle the FR3 robot despite the above-mentioned challenges. An example of stacking using the Franka arm is shown in Fig. \ref{real_stack}. This real-world study is limited to a single task and executes the best simulated trajectory rather than replanning online; broader multi-trial evaluation is left for future work.

\section{Conclusion and Future Works}

In this paper, we introduced iCEM+TL, a framework that extends iCEM with task decomposition and transfer mechanisms for complex robotic manipulation. By integrating TL and RR into the optimization process, the method enables knowledge reuse and structured trajectory search across tasks. We further analyze upstream task selection based on elite trajectory consistency under downstream rewards, providing an interpretable criterion for transfer effectiveness in optimization-based planners. Experiments across benchmark environments, including FetchStack, FetchSlide, and Shelf, demonstrate improvements of up to $23\%$ over the baselines, while real-world experiments on a Franka Emika robot confirm the feasibility of the approach.
Our results provide practical insights into applying TL within optimization-based planners. In particular, performance depends on balancing exploration and exploitation: larger sample sizes improve exploration, while a compact elite set preserves strong guidance from high-quality trajectories, including transferred ones. In addition, RR is essential for enabling complex tasks, and appropriate planning horizons are required for effective optimization. Overall, these findings show that incorporating task decomposition into optimization-based planners can substantially improve performance without long-horizon policy training, offering a more interpretable and computationally efficient alternative to purely learning-based approaches.

Future work will extend the framework to more challenging and realistic manipulation settings. One direction is to integrate iCEM+TL with high-level learned models for hierarchical planning, where learned policies provide coarse guidance and the planner refines low-level trajectories. We also plan to study more complex real-world settings involving orientation constraints and contact-rich scenarios, which introduce challenges such as contact discontinuities and sensitivity to pose errors. Finally, we aim to investigate adaptive transfer strategies to improve robustness and generalization across diverse tasks.

\bibliographystyle{unsrt}
\bibliography{References}

\end{document}